%% file: main.tex
\documentclass[10pt,twocolumn,letterpaper]{article}
\PassOptionsToPackage{table,dvipsnames}{xcolor}
\PassOptionsToPackage{nameinlink}{cleveref}

\usepackage[pagenumbers]{cvpr} 
\input{preamble}
\input{shortcut}
\definecolor{cvprblue}{rgb}{0.21,0.49,0.74}
\usepackage[pagebackref,breaklinks,colorlinks,allcolors=cvprblue]{hyperref}

\def\confName{CVPR}
\def\confYear{2027}

\title{\acs*{RetrACER}: Multi-Session Navigation Memory for \\Efficient Lifelong Spatio-Temporal Reasoning}

\author{
    Bokeon Suh\textsuperscript{1}\:
    Yumin Lee\textsuperscript{1}\:
    Giseop Kim\textsuperscript{1}\setcounter{footnote}{1}\thanks{Corresponding author} \\
    \textsuperscript{1}Department of Robotics and Mechatronics Engineering, DGIST, Republic of Korea \\
    {\tt\small \{bksuh, yumin.lee, gsk\}@dgist.ac.kr}
}

\begin{document}
\raggedbottom
\maketitle
\input{sec/abstract}
\acresetall
\input{sec/introduction}
\input{sec/related_work}

\input{sec/method}
\input{sec/experiments}
\input{sec/limitations}
\input{sec/conclusion}

\section*{LLM Usage Statement}
We used ChatGPT to edit the text of this paper.
Its use was limited to grammar correction, rephrasing of sentences that we had written, and improving readability.\\
{
    \small
    \bibliographystyle{ieeenat_fullname}
    \bibliography{main}
}
\input{sec/supplementary}

\end{document}

%% file: preamble.tex
\usepackage{makecell}
\usepackage{multirow}
\usepackage{pifont}     
\usepackage{acro}
\usepackage{siunitx}
\usepackage{microtype}
\usepackage{longtable,tabularx,float}
\AtEndPreamble{%
    \crefname{table}{Table}{Tables}%
    \Crefname{table}{Table}{Tables}%
}

%% file: shortcut.tex
\newcommand{\ours}{\ac{RetrACER}\xspace}
\newcommand{\ourmem}{\ac{PAST-Mem}\xspace}
\newcommand{\oursdata}{\ac{MS-NaVQA}\xspace}
\newcommand{\retracerinitial}[1]{#1}
\newcommand{\retracershort}[1]{#1}

\newcommand{\remembr}{ReMEmbR\xspace}
\newcommand{\navqa}{NaVQA\xspace}
\newcommand{\ocnavqa}{OC-NaVQA\xspace}
\newcommand{\daaam}{DAAAM\xspace}
\newcommand{\starmem}{STaR\xspace}
\newcommand{\conceptgraphs}{ConceptGraphs\xspace}
\newcommand{\sota}{state-of-the-art\xspace}
\newcommand{\ltmem}{LT-Mem\xspace}

\newcommand{\cmark}{\textcolor{green!60!black}{\ding{51}}}
\newcommand{\xmark}{\textcolor{red!75!black}{\ding{55}}}

\newcommand{\supText}{\mathrm{text}}
\newcommand{\supVis}{\mathrm{vis}}
\newcommand{\subObj}{\mathrm{obj}}
\newcommand{\subCue}{\mathrm{cue}}
\newcommand{\subCrop}{\mathrm{crop}}

\DeclareMathOperator{\UpdateMemory}{\mathtt{UpdateMemory}}
\DeclareMathOperator{\BuildEvidence}{\mathtt{BuildEvidence}}
\DeclareMathOperator{\GenerateAnswer}{\mathtt{GenerateAnswer}}

\definecolor{lightergray}{gray}{0.96}
\newcolumntype{g}{>{\columncolor{lightergray}}c}

\definecolor{first}{HTML}{C0E5CD}
\definecolor{second}{HTML}{E4EEBE}
\definecolor{third}{HTML}{FEF8C7}

\newcommand{\fst}[1]{\cellcolor{first}\textbf{#1}}
\newcommand{\snd}[1]{\cellcolor{second}#1}
\newcommand{\trd}[1]{\cellcolor{third}#1}

\DeclareAcronym{RetrACER}{
  short = \retracershort{RetrACER},
  long  = \retracerinitial{Retr}ieval-\retracerinitial{A}ugmented \retracerinitial{C}ross-session \retracerinitial{E}fficient Embodied \retracerinitial{R}easoning
}
\DeclareAcronym{PAST-Mem}{
  short = PAST-Mem,
  long  = Place-Addressed Spatio-Temporal Memory
}
\DeclareAcronym{MS-NaVQA}{
  short = MS-NaVQA,
  long  = Multi-Session Navigation Question Answering
}
\DeclareAcronym{LLM}{
  short = LLM,
  long  = Large Language Model
}
\DeclareAcronym{RAG}{
  short = RAG,
  long  = Retrieval-Augmented Generation
}
\DeclareAcronym{SG}{
  short = SG,
  long  = Scene Graph
}
\DeclareAcronym{VLM}{
  short = VLM,
  long  = Vision-Language Model
}
\DeclareAcronym{QA}{
  short = QA,
  long  = question answering
}

%% file: sec/abstract.tex
\begin{abstract}
Robots that revisit an environment accumulate experiences that help answer questions about its past, such as whether a place is more crowded than on a previous visit or whether the robot used a different building entrance.
Such questions require retaining observations across visits and efficiently retrieving evidence for comparison.
We introduce {\renewcommand{\retracerinitial}[1]{\textbf{\underline{#1}}}\renewcommand{\retracershort}[1]{\textbf{#1}}\ac{RetrACER}}, a multi-session navigation memory system for efficient lifelong spatio-temporal reasoning.
\ac{PAST-Mem} provides place-level memory scoping through a spatial index decoupled from detailed observations, bounding index entries per place while preserving evidence across visits.
At the record level, Type-Intent Evidence Plans use query type and intent to select and process records through semantic, geometric, and temporal operations, avoiding iterative \ac{LLM}-directed tool calls.
To evaluate reasoning across long-term experiences, we introduce \ac{MS-NaVQA}, a benchmark covering historical observations, events, and comparisons.
With a locally hosted 9B model, \mbox{\ours} achieves the highest question accuracy among the evaluated methods on both single-session benchmarks and \oursdata, while reducing query latency and token usage.
\end{abstract}

%% file: sec/introduction.tex
\input{figure/fig_memory_motivation.tex}
\section{Introduction}
\label{sec:intro}
Recent advances in \acp{LLM} and \acp{VLM} have enabled embodied agents to use past observations during extended navigation~\cite{gorlo2026describe,anwar2025remembr,mao2025meta}, exploration~\cite{yang20253dmem,wang2026explore,jiang2024roboexp}, and manipulation~\cite{sridhar2026scaling,liu2025dynamem,shi2026memoryvla}.
As observations accumulate, their history can exceed practical model context limits.
Embodied memory allows agents to retain past experience and retrieve relevant information when needed, following the principle of \ac{RAG}~\cite{lewis2020retrieval}.
Existing navigation memories take two broad forms: observation-indexed memory and 3D \ac{SG} memory.
Observation-indexed memory~\cite{anwar2025remembr,hu2026raven,mao2025meta} stores temporal observation segments with semantic and spatio-temporal metadata, including poses and timestamps.
\Ac{SG} memory, in contrast, organizes observations into persistent objects and their relations~\cite{gorlo2026describe,gu2024conceptgraphs,maggio2024clio,werby2024hierarchical}.
Both have shown promise on long-horizon navigation \ac{QA}, but predominantly within a single session.

As experience accumulates across sessions, navigation memory must preserve answer accuracy while efficiently retrieving and constructing \emph{evidence}, the question-relevant information used to answer a query.
Questions about environmental changes require distinguishing observations from different visits (\cref{fig:memory_motivation}, top).
Observation-indexed memories append a searchable record per observation segment, so the retrieval pool grows with each revisit and earlier records can compete with near-duplicates under a fixed retrieval budget.
For \ac{SG} memories, consolidating observations into persistent object representations must preserve historical states for cross-session comparison.
The challenge is to preserve historical evidence while scoping retrieval to query-relevant observations.

Existing methods~\cite{anwar2025remembr,gorlo2026describe,hu2026raven,mao2025meta} use iterative \ac{LLM}/\ac{VLM}-directed tool calls to retrieve and process evidence, which can increase query latency and token usage~\cite{xu2023rewoo,kim2024llm}.
Yet a query itself provides cues for retrieval and evidence construction in advance through its query type (the expected answer category) and intent (the specific information sought).

To this end, we jointly design memory organization and evidence construction for multi-session navigation in \ac{RetrACER}.
Its \ac{PAST-Mem} separates detailed object records and frame records from a place retrieval index that links a bounded number of embeddings per \emph{place} to its grid-cell identifier.
Here, \emph{records} refer to individual frame or object entries with associated features and metadata.
On revisits, \ours links new observations to existing place addresses, retaining earlier records (\cref{fig:memory_motivation}, bottom).
This bounds index size by spatial coverage while preserving observation-specific spatial and temporal information.
When queried, \ourmem's index narrows the search to relevant places (\emph{place-level memory scoping}), while \ours selects a predefined Type-Intent Evidence Plan, which applies semantic, geometric, and temporal operations to the scoped records without iterative tool calls (\emph{intent-guided evidence construction}).
\ours achieves the highest question accuracy among the evaluated methods on both single-session benchmarks and our \ac{MS-NaVQA} dataset, with improved query efficiency and fewer indexed entries than the baselines.

In summary, our work makes the following contributions:
\begin{itemize}[leftmargin=*, nosep, topsep=2pt, partopsep=0pt, parsep=0pt]
    \item Our \textbf{\ours} combines place-level memory scoping with record-level evidence construction to support accurate, efficient lifelong spatio-temporal question answering across repeated deployments.
    \begin{itemize}[label=$\circ$, leftmargin=0.8em, labelsep=0.3em, noitemsep, topsep=0pt, partopsep=0pt]
        \item \textbf{\ourmem} uses shared place addresses to organize Place Synopses, Scene Codebooks, a Frame Archive, and an Object Catalog, bounding index entries per place while retaining detailed evidence.
        \item \textbf{Type-Intent Evidence Plans} transform retained records into question-specific evidence using semantic, geometric, and temporal operations.
    \end{itemize}
    \item Finally, we introduce \textbf{\oursdata}, a multi-session navigation \ac{QA} dataset for reasoning over experiences accumulated across repeated traversals, including comparisons between visits.
\end{itemize}

%% file: figure/fig_memory_motivation.tex
\begin{figure}[t]
    \centering
    \includegraphics[width=0.9\linewidth]{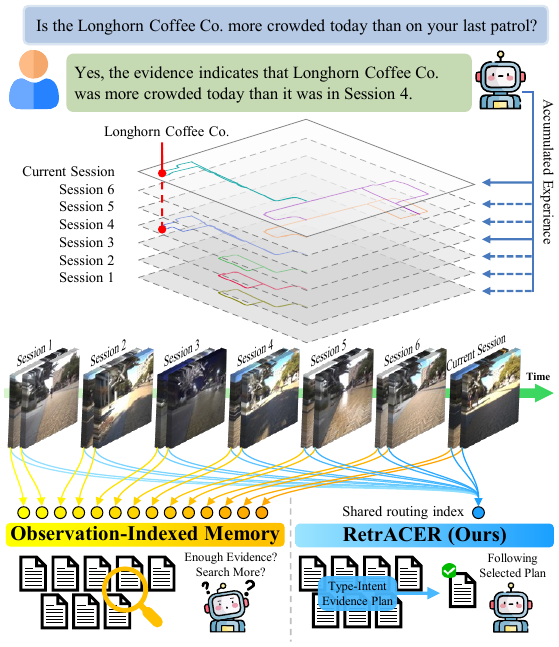}
    \caption{\emph{Top:} A multi-session navigation question requiring comparison of observations across visits.
    \emph{Bottom:} Repeated visits add observation-level index entries (orange dots), whereas \acs*{RetrACER} links observations to a reusable place address (blue dot).}
    \label{fig:memory_motivation}
    \vspace{-3mm}
\end{figure}

%% file: sec/related_work.tex
\section{Related Work}
\label{sec:rw}
Prior work on embodied memory addresses how to represent past experience, organize retrieval, and acquire evidence for answering.

\subsection{Navigation Memory for Embodied Agents}
\label{sec:rw_navigation_memory}
Navigation memory supports \acp{LLM} and \acp{VLM} in long-horizon operations through descriptive summaries, visual embeddings, or 3D \acp{SG}.
One approach, \textbf{Observation-Indexed Memory}, organizes individual frames or temporal segments as independently searchable records with semantic, spatial, and temporal metadata.
\emph{Scene-level descriptions} refer to stored frame captions or scene and region summaries.
Frame- or segment-level captions~\cite{anwar2025remembr,mao2025meta} can be supplemented by visual embeddings and keyframes for verification~\cite{hu2026raven,mao2025meta}.
Another approach, \textbf{\ac{SG} Memory}, organizes observations into persistent objects and their relations~\cite{gu2024conceptgraphs,maggio2024clio,werby2024hierarchical}.
Individually addressable object records with semantic attributes and 3D geometry provide \emph{object-centric memory} for spatial reasoning and retrieval.
Beyond individual objects, \daaam~\cite{gorlo2026describe} stores region summaries as scene-level descriptions.
\starmem~\cite{yuan2026star} combines object-centric memory with scene-level descriptions.
\emph{Multi-session QA evaluation} requires answering questions using records from independently acquired sessions in the original paper; untested extensions are marked -- in \cref{tab:memory_comparison}.
\emph{Bounded index growth} means that the retrieval-index size remains bounded as observations accumulate within fixed spatial coverage.
With consistent spatial and temporal references, observation-indexed memories can append records across sessions, but their retrieval pools grow.
\ac{SG} memories must additionally resolve cross-session object correspondences.
\ltmem~\cite{lee2026lt} maintains cross-session object identities and state histories and evaluates cross-session questions about designated object instances.
\Cref{tab:memory_comparison} compares the capabilities of representative embodied memory systems.
\input{table/tab_memory_comparison}
\input{figure/fig_method_overview.tex}

\subsection{From Text-Based \acs*{RAG} to Embodied Memory}
\label{sec:rw_embodied_rag}
Studies on \ac{RAG} and memory in NLP have explored hierarchical organization~\cite{chen2026improving,lu2026hichunk,sarthi2024raptor,zhao2026hierarchical}, coarse-to-fine retrieval~\cite{shin2026hikey,zhao2025funnelrag}, query-aware compression~\cite{louis2026oscar,xu2024recomp}, and semantic consolidation~\cite{liu2026simplemem} to improve efficiency.
Document units, from paragraphs and sections to entire documents, support hierarchical retrieval.
However, organizing embodied observations for retrieval is nontrivial: content, position, time, and viewpoint define relationships that need not align with a single hierarchy.
Compact retrieval representations may omit temporal and geometric distinctions needed for answering.
We adapt these ideas using persistent \emph{places} to organize navigation observations.
\ourmem achieves bounded index growth by limiting entries per persistent \emph{place} (\cref{eq:routing_bound}), while allowing detailed object records and frame records to grow.

\subsection{Agentic Retrieval and Evidence Construction}
\label{sec:rw_agentic_retrieval}
Agentic approaches use \acp{LLM}/\acp{VLM} to select tools and acquire information through external calls~\cite{yao2023react,anwar2025remembr,gorlo2026describe,hu2026raven}.
When retrieval is iterative, the model repeatedly decides what to retrieve next and whether the retrieved information is sufficient to answer the query.
Even when these calls retrieve relevant evidence, semantically related but irrelevant records may be included in the context provided to the model and mislead it during answer generation~\cite{amiraz2025distracting,jin2025long}.
The position of relevant evidence can also affect its use in long contexts~\cite{liu2024lost}.
Beyond retrieval, some work delegates arithmetic to external tools~\cite{schick2023toolformer}, logical inference to symbolic solvers~\cite{olausson2023linc,ye2023satlm}, and spatial operations to dedicated modules~\cite{yuan2024visual}.
Building on this division of labor, \ours selects a predefined Type-Intent Evidence Plan for intent-guided evidence construction using the query type and intent parsed by the \ac{LLM}.
The plan specifies which records to retrieve and which semantic, geometric, and temporal operations to apply, constructing structured, query-conditioned evidence without repeated language-model orchestration.
The language model then answers the question using this evidence.

%% file: table/tab_memory_comparison.tex
\begin{table}[t]
\centering
\caption{
Comparison under the criteria in \cref{sec:rw}.
\cmark: supported; \xmark: unsupported by the described design;
--: potentially feasible, but neither targeted nor demonstrated in the original paper.
}
\label{tab:memory_comparison}
\setlength{\tabcolsep}{1.5pt}
\renewcommand{\arraystretch}{1.15}
\resizebox{\columnwidth}{!}{%
\begin{tabular}{llcccc}
\toprule
\makecell[l]{\textbf{Venue}} &
\makecell[l]{\textbf{Method}} &
\makecell[c]{\textbf{Object-Centric}\\\textbf{Memory}} &
\makecell[c]{\textbf{Scene-Level}\\\textbf{Descriptions}} &
\makecell[c]{\textbf{Multi-Session}\\\textbf{QA Evaluation}} &
\makecell[c]{\textbf{Bounded}\\\textbf{Index Growth}} \\
\midrule
ICRA~'24 & \conceptgraphs~\cite{gu2024conceptgraphs}
  & \cmark & \xmark & -- & -- \\
ICRA~'25 & \remembr~\cite{anwar2025remembr}
  & \xmark & \cmark & -- & \xmark \\
RA-L~'26 & \starmem~\cite{yuan2026star}
  & \cmark & \cmark & -- & \xmark \\
CVPR~'26 & \daaam~\cite{gorlo2026describe}
  & \cmark & \cmark & -- & -- \\
IROS~'26 & \ltmem~\cite{lee2026lt}
  & \cmark & \xmark & \cmark & -- \\
\midrule
{} & \textbf{\acs*{RetrACER} (Ours)}
  & \cmark & \cmark & \cmark & \cmark \\
\bottomrule
\end{tabular}%
}
\end{table}

%% file: figure/fig_method_overview.tex
\begin{figure*}[t]
    \centering
    \includegraphics[width=1\textwidth]{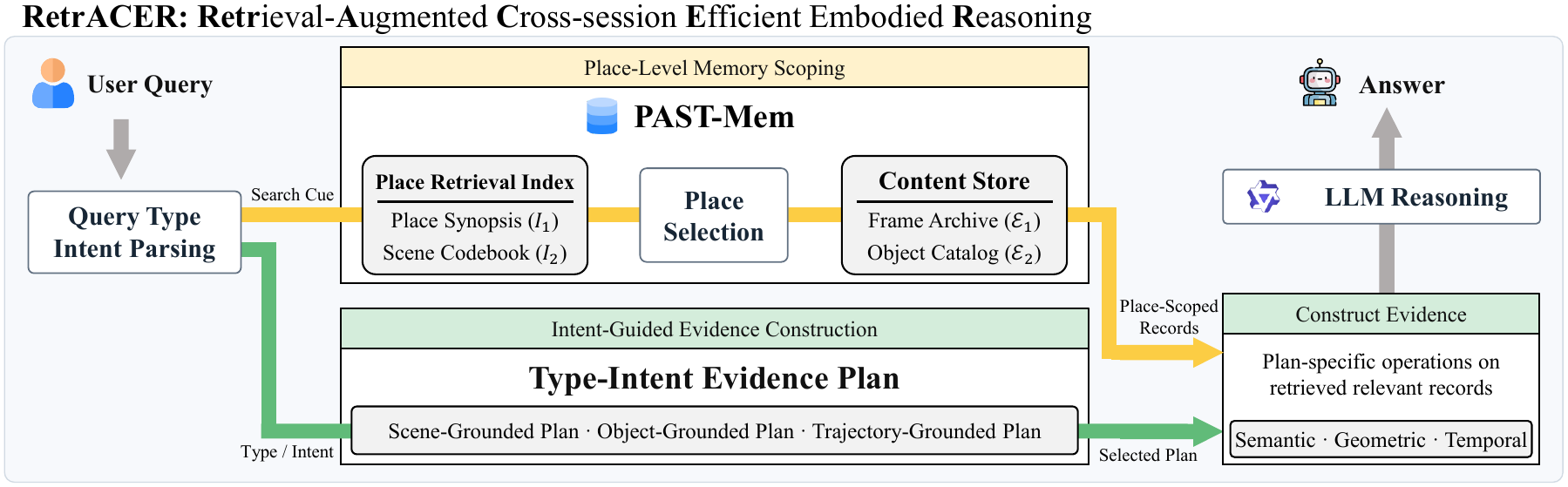}
    \caption{
    \textbf{Overview of \acs*{RetrACER}.}
    \emph{Place-level memory scoping} selects relevant \emph{places} in \acs*{PAST-Mem} using search cues.
    \emph{Intent-guided evidence construction} selects a Type-Intent Evidence Plan using query type and intent.
    The paths meet at record-level evidence construction, where plan-specific selection and semantic, geometric, and temporal operations transform place-scoped records into answer evidence.
    }
    \label{fig:method_overview}
    \vspace{-3mm}
\end{figure*}

%% file: sec/method.tex
\section{Approach}
\label{sec:approach}
We first formalize our task in terms of incremental memory updates and query-time evidence retrieval (\cref{sec:problem_formulation}).
As shown in \cref{fig:method_overview}, \ours combines \emph{place-level memory scoping} through \textbf{\ourmem} (\cref{sec:memory_building}) with \emph{intent-guided evidence construction} at the record level through \textbf{Type-Intent Evidence Plans} (\cref{sec:query_answering}).

\subsection{Problem Formulation}
\label{sec:problem_formulation}
We consider an embodied agent that operates in an environment over one or more sessions and later answers natural-language questions based on its accumulated observations.
During each session $s$, the agent traverses the environment and gathers a sequence of posed RGB-D observations $\mathcal{Z}_s=\{z_i\}_{i=1}^{N_s}$, where $N_s$ is the observation count and $i$ indexes frames within session $s$.
Each observation $z_i=(I_i, D_i, \mathbf{p}_i, \theta_i, \tau_i)$ contains an RGB image, a depth image, the agent's position and heading, and a timestamp.
We assume that agent poses are provided in a common global coordinate frame across sessions.
Such alignment can be obtained using existing collaborative mapping systems, such as MR.ScaleMaster~\cite{ju2026mrscalemaster}.
The agent maintains a persistent memory that is updated incrementally:
\begin{equation}
    \mathcal{M}_{1:s} = \UpdateMemory(\mathcal{M}_{1:s-1}, \mathcal{Z}_s),
    \quad
    \mathcal{M}_{0} = \emptyset.
    \label{eq:memory_update}
\end{equation}

At query time, the agent answers a natural-language question $q$ using the accumulated memory $\mathcal{M}_{1:s}$.
Providing the full memory to a reasoner (e.g., a large language model) can exceed its context-window capacity.
We define \emph{evidence}, $\mathcal{W}_q$, as question-relevant information selected or computed from stored records and supplied to the reasoner:
\begin{equation}
    \mathcal{W}_q = \BuildEvidence(q,\mathcal{M}_{1:s}).
    \label{eq:retrieve}
\end{equation}
We maximize expected answer accuracy:
\begin{equation}
    \max_{\substack{\UpdateMemory,\\\BuildEvidence}}\,
    \mathbb{E}_{q}
    [\mathrm{acc}(\GenerateAnswer(q,\mathcal{W}_q),y^\star)],
    \label{eq:problem_objective}
\end{equation}
where $\GenerateAnswer$ produces an answer from $q$ and $\mathcal{W}_q$, and $y^\star$ denotes the ground-truth answer.
We also seek to keep evidence retrieval efficient as observations accumulate.

\subsection{\texorpdfstring{\acs*{RetrACER}}{RetrACER} Memory}
\label{sec:memory_building}
The memory of \ours, termed \ourmem, links place embeddings to grid-cell identifiers in a \emph{place retrieval index} for selecting query-relevant places.
We bound index entries on revisits using persistent spatial addresses shared across sessions.
We index observations by the horizontal position of their $\mathrm{SE}(3)$ poses, using fixed square cells of width $\ell$ to define \emph{places}.
Let $g(\mathbf{p})$ denote the grid-cell identifier for position $\mathbf{p}$, and $\mathcal{G}$ the set of visited cell identifiers.

Observations assigned to the same \emph{place} may differ in acquisition time, viewpoint, and observed content.
These observations share the \emph{place}'s spatial address for retrieval, while their records remain separately accessible.
The index $\mathcal{I}$ provides place-level memory scoping using the Place Synopsis ($\mathcal{I}_1$) and Scene Codebook ($\mathcal{I}_2$).
The \emph{content store} $\mathcal{E}$ retains detailed frame records and object records in the Frame Archive ($\mathcal{E}_1$) and Object Catalog ($\mathcal{E}_2$).
Writing $\mathcal{M}$ for $\mathcal{M}_{1:s}$ hereafter, we express our Place-Addressed Spatio-Temporal Memory (\ourmem) as
\begin{equation}
    \mathcal{M}
    = \underbrace{\mathcal{I}}_{\mathcal{I}_1 \cup \mathcal{I}_2}
    \cup \underbrace{\mathcal{E}}_{\mathcal{E}_1 \cup \mathcal{E}_2}.
    \label{eq:memory_split}
\end{equation}
The following subsections detail $\mathcal{I}_1$, $\mathcal{I}_2$, $\mathcal{E}_1$, and $\mathcal{E}_2$.

\subsubsection{Place Retrieval Index}
\label{sec:sub_routing_index}
Two complementary signals select query-relevant \emph{places}.

\textbf{Place Synopsis ($\mathcal{I}_1$).}
A synopsis provides a compact description of a \emph{place}.
An \ac{LLM} updates the synopsis using frame captions grouped within the same \emph{place}, prioritizing content supported by repeated observations across sessions.
The resulting synopsis, when nonempty, is embedded as a single index entry.

\textbf{Scene Codebook ($\mathcal{I}_2$).}
A scene codebook is a bounded set of representative vectors obtained from a \emph{place}'s caption embeddings.
It provides retrieval cues for observed content that may be absent from the place synopsis.
Let $n_g$ count observations at \emph{place} $g$ and $K$ bound the number of codebook vectors.
We store all normalized caption embeddings when $n_g \le K$; otherwise, spherical $k$-means yields $K$ unit-length cluster centroids.

Because each \emph{place} contributes at most one synopsis entry and at most $K$ codebook entries, the index size is bounded by the number of visited \emph{places}:
\begin{equation}
    |\mathcal{I}|
    = |\mathcal{I}_1| + |\mathcal{I}_2|
    \le (1+K)|\mathcal{G}|.
    \label{eq:routing_bound}
\end{equation}
For fixed cell width $\ell$ and codebook capacity $K$, this bound depends on spatial coverage rather than observation count, while detailed records can accumulate across revisits.

\subsubsection{Content Store}
\label{sec:sub_content_store}
A \emph{record} is a structured entry for an individual frame or object.
Frame records store captions, embeddings, poses, and timestamps; object records store embeddings, 3D positions, and links to supporting frames.
Frame identifiers (\texttt{FrameID}s) link the two record types.

\textbf{Frame Archive ($\mathcal{E}_1$).}
For frame $I_i$, $\mathcal{E}_1$ stores an NVILA caption~\cite{liu2025nvila}, its text embedding $\mathbf{e}^{\supText}_{I_i}$, the image embedding $\mathbf{e}^{\supVis}_{I_i}$, and $(\mathbf{p}_i,\theta_i,\tau_i)$.
Here, superscripts specify the embedding space: $\supVis$ for SigLIP 2~\cite{tschannen2025siglip} and $\supText$ for mxbai-embed-large-v1~\cite{embed2024mxbai}; subscripts specify the represented entity.
Each record has a \texttt{FrameID} and is addressed by $g(\mathbf{p}_i)$.
Records from different visits remain separate after index updates.

\textbf{Object Catalog ($\mathcal{E}_2$).}
Each object record stores a representative crop embedding $\mathbf{e}^{\supVis}_{\subObj}$, a mean global 3D position $\mathbf{p}_{\subObj}$, and supporting \texttt{FrameID}s.
For each retained SAM~3~\cite{carion2026sam3segmentconcepts} mask from $I_i$, we extract a crop embedding $\mathbf{e}^{\supVis}_{\subCrop}$ and back-project aligned depth to a global 3D position.
We associate the mask with a nearby record if $\cos(\mathbf{e}^{\supVis}_{\subCrop},\mathbf{e}^{\supVis}_{\subObj})$ meets a threshold; otherwise, we initialize a new record from this observation.
For a matched mask, we update the record's mean position and add the corresponding \texttt{FrameID}.
We replace $\mathbf{e}^{\supVis}_{\subObj}$ with $\mathbf{e}^{\supVis}_{\subCrop}$ if the latter has higher cosine similarity to the running mean of associated crop embeddings than the current representative.
We concatenate and encode selected captions from the record's current supporting frames to obtain $\mathbf{e}^{\supText}_{\subObj}$.

Object records are addressed by $g(\mathbf{p}_{\subObj})$, while frame records are addressed by $g(\mathbf{p}_i)$.
Shared \texttt{FrameID}s link object records to supporting frame records in both directions, even when the object and agent occupy different cells.
\subsection{\texorpdfstring{\acs*{RetrACER}}{RetrACER} Inference}
\label{sec:query_answering}
At query time, \ours combines place-level memory scoping via $\mathcal{I}$ with intent-guided evidence construction over records in $\mathcal{E}$ to produce $\mathcal{W}_q$ (\cref{eq:retrieve}).
\subsubsection{Place-Level Memory Scoping}
A \emph{search cue} is a textual description taken from the question or generated by the \ac{LLM} to guide retrieval.
We encode each cue as $\mathbf{e}^{\supText}_{\subCue}$ in the text embedding space.
We select \emph{places} by cosine similarity between $\mathbf{e}^{\supText}_{\subCue}$ and their synopsis embeddings in $\mathcal{I}_1$.
If synopsis retrieval is empty, or its best match scores below a threshold, we instead retrieve the top-ranked codebook entries in $\mathcal{I}_2$ and include the $3\times3$ neighborhoods of their associated cells.
The selected \emph{places} scope access to frame records in $\mathcal{E}_1$ and object records in $\mathcal{E}_2$.

\subsubsection{Intent-Guided Evidence Construction}
\textbf{Query Type and Intent Parsing.}
\emph{Query type} specifies the answer category (text, binary, position, time, or duration), as in \remembr~\cite{anwar2025remembr}.
\emph{Intent} specifies the information sought within that category.
For example, ``Where is the bike rack closest to the yellow police pole?'' and ``Where did you last leave the building?'' are both position queries, but require object localization with spatial reasoning and trajectory-based exit detection, respectively.
Three predefined Type-Intent Evidence Plans (\cref{tab:evidence_plans}) cover the query types and intents evaluated here.
An \ac{LLM} identifies the query type and fine-grained intent to select a plan.
The selected plan constructs question-specific evidence $\mathcal{W}_q$ through predefined semantic, geometric, and temporal operations on place-scoped records, without an iterative loop in which the \ac{LLM} chooses subsequent tool calls from intermediate results.

\textbf{Scene-Grounded Plan.}
This plan answers descriptive, action, relation, and ego-state questions using retrieved frame records.
SigLIP 2's text encoder maps target descriptions to $\mathbf{e}^{\supVis}_{\subCue}$ for visual matching.
Within selected \emph{places}, we retrieve frames in $\mathcal{E}_1$ by matching the question's $\mathbf{e}^{\supText}_{\subCue}$ to $\mathbf{e}^{\supText}_{I_i}$ and a target's $\mathbf{e}^{\supVis}_{\subCue}$ to $\mathbf{e}^{\supVis}_{I_i}$.
Agent positions and timestamps distinguish observations across places and visits.
Ego-state evidence additionally includes counts and timestamp spans of spatially relevant frames.

\textbf{Object-Grounded Plan.}
Existence queries start from frames in $\mathcal{E}_1$ and follow \texttt{FrameID}s to associated objects in $\mathcal{E}_2$.
An \ac{LLM} judges whether the target was observed using similarities between $\mathbf{e}^{\supVis}_{\subCue}$ and $\mathbf{e}^{\supVis}_{\subObj}$, counts of supporting frames above a similarity threshold, and dated captions.
Localization queries instead retrieve object candidates in $\mathcal{E}_2$ within selected \emph{places}, matching $\mathbf{e}^{\supVis}_{\subCue}$ to $\mathbf{e}^{\supVis}_{\subObj}$ and $\mathbf{e}^{\supText}_{\subCue}$ to $\mathbf{e}^{\supText}_{\subObj}$.
\ac{LLM}-generated descriptions broaden retrieval; we group nearby candidates and access supporting frames using \texttt{FrameID}s.
Distances to the agent or reference objects encode spatial constraints (\eg, proximity to the yellow police pole in our earlier example).
The \ac{LLM} checks supporting captions to select a candidate.
The output pairs $\mathbf{p}_{\subObj}$ with the nearest supporting viewpoint in the ground plane on the observed trajectory.

\textbf{Trajectory-Grounded Plan.}
Time, duration, and event-location queries use timestamp-ordered frames from selected \emph{places}.
The \ac{LLM} specifies the visual or motion signal and target event occurrence.
We detect event points using similarity differences, peaks, or motion thresholds and return their timestamps or recorded agent positions; durations are the differences between endpoint timestamps.

These plans provide practical coverage of the evaluated queries without claiming an exhaustive or mutually exclusive taxonomy.
With \ourmem, the system outperforms \remembr and our \daaam adaptation in multi-session question accuracy and query latency (\cref{sec:experimental_settings,tab:msnavqa_results}).

%% file: sec/experiments.tex
\section{Experiments}
\label{sec:experiments}
\subsection{Experimental Settings}
\label{sec:experimental_settings}
\textbf{Implementation Details.}
We use an RTX 4090 workstation with \qty{64}{GB} RAM and serve Qwen3.5-9B via Ollama on NVIDIA DGX Spark.
Object masks were obtained by SAM~3 using automatic point-grid prompts, captions were generated by NVILA-Lite-8B, and Milvus supports vector retrieval.
We set $\ell=\qty{10}{\meter}$, $K=8$, and the synopsis-retrieval similarity threshold to $0.65$.
Object association uses a \qty{2}{\meter} radius and a cosine-similarity threshold of $0.80$.

\input{table/tab_evidence_plans.tex}
\input{table/tab_navqa_results.tex}
\textbf{Datasets.}
We evaluate on the single-session benchmarks \navqa~\cite{anwar2025remembr} and \ocnavqa~\cite{gorlo2026describe}, and our multi-session benchmark \oursdata.
\navqa contains 210 questions on indoor and outdoor campus traversals from CODa~\cite{zhang2024robust}.
Its \emph{Short}, \emph{Medium}, and \emph{Long} memory horizons cover up to \qty{23.5}{\minute} and \qty{1.04}{\kilo\meter}.
However, some questions refer to observations outside their assigned memory windows, and position labels specify where the robot observed the target rather than the target's 3D position.
\ocnavqa addresses these limitations by using observation histories up to each question and object-position labels, with histories spanning up to \qty{35.8}{\minute} and \qty{1.64}{\kilo\meter}.

Still, neither benchmark requires observations from earlier sessions.
We therefore introduce \oursdata to evaluate \ac{QA} using memory accumulated across sessions in CODa, spanning up to \qty{164.4}{\minute} and \qty{7.6}{\kilo\meter} for the longest question.
Its 70 \ac{QA} pairs fall into three categories (\cref{fig:question_categories}).
\emph{Historical Verification} checks changes in observed conditions, such as whether a lobby is more crowded than on a previous visit.
\emph{Historical Comparison} compares events or states across sessions, such as whether the robot used a different entrance on an earlier visit.
\emph{Historical Occurrence} asks about first encounters with objects or places.

\textbf{Baselines.}
We compare \ours with recent baselines, including the \sota method \daaam~\cite{gorlo2026describe}, on the single-session benchmarks.
For \oursdata, we compare with the observation-indexed memory \remembr~\cite{anwar2025remembr}, whose records are accumulated across sessions, and the \ac{SG} memory \daaam.
We modify \daaam to support multi-session evaluation by combining session graphs while preserving session identifiers and timestamps.
Note that this extension retains the original retrieval and reasoning design without introducing a new cross-session object association policy.
We also include Majority Pick, which always predicts ``yes,'' the majority label in \oursdata.
We excluded \ltmem~\cite{lee2026lt} because it requires target objects to be specified before memory construction, whereas neither questions nor target objects are provided in advance.

\input{figure/fig_question_categories.tex}
\textbf{Single-Session Results.}
We report the performance of \ours and the baselines on \navqa and \ocnavqa in \cref{tab:navqa_results,tab:ocnavqa_results}, respectively.
Using a locally hosted 9B language model, \ours outperforms all baselines, including GPT-based ones, in question accuracy and overall positional error on both benchmarks.
However, on \ocnavqa, we observed cases in which the \ac{LLM} failed to resolve the intended event occurrence when the query provided no explicit cue (\eg, ``When did you cross the street?''), leading the Trajectory-Grounded Plan to make incorrect selections among the additional competing candidates in longer observation histories and produce large errors that increased the mean temporal error.

\textbf{Results on \oursdata.}
\input{table/tab_ocnavqa_results.tex}
As shown in \cref{tab:msnavqa_results}, \ours achieves the highest question accuracy and balanced accuracy, which accounts for label imbalance, with the lowest average query latency.
We observe cases in which \remembr and \daaam answer incorrectly, stating that they lack sufficient evidence for cross-session comparisons.
In contrast, \ours retrieves observations from both visits in these cases, as shared \emph{place} addresses keep records from repeated visits within a common retrieval scope.
In addition, while \remembr and \daaam rely on iterative \ac{LLM}-directed tool calls, \ours constructs evidence through the selected plan, which reduces the retrieval decisions and computations delegated to the \ac{LLM}, supporting effective and efficient reasoning over accumulated navigation experience.

\input{table/tab_msnavqa_results.tex}
\input{figure/fig_index_growth.tex}
\textbf{Retrieval Index Growth.}
As shown in \cref{fig:index_growth}, \remembr appends observation records at $\rho=0.33$ records/s, so its retrieval index grows with operation time.
In contrast, \ours updates existing entries in $\mathcal{I}_1$ and $\mathcal{I}_2$ when the agent follows trajectories similar to those of earlier sessions, adding only 76 entries across sessions 2--3, versus 758 for \remembr.
Although initially larger, the retrieval index of \ours reaches 2,253 entries versus 3,253 for \remembr after seven sessions, a 30.7\% reduction under the evaluated settings and trajectories.
Thus, the retrieval-index bound depends on spatial coverage for fixed $\ell$ and $K$ (\cref{eq:routing_bound}); the growing content store is excluded from these counts and remains searchable within selected \emph{places}.

\textbf{Memory Density and Retrieval Budget.}
Since our method stores observations more frequently than \remembr, we first examine whether increasing the memory density of \remembr improves its performance.
We increase its memory density $\rho$ from the default of approximately 0.33 to 10 records/s to match \ours, and additionally raise the retrieval budget from 5 to 150 entries per tool call to reflect the 30-fold increase in memory density.

As shown in \cref{tab:density_budget}, increasing the memory density improves performance in some settings.
Further increasing the retrieval budget, however, substantially raises average token usage per question (input and output) and query latency without consistently improving performance.
Meanwhile, \ours applies query-specific spatial and temporal operations to the retrieved records before passing evidence to the \ac{LLM} and achieves lower latency and token usage across all three benchmarks.
The results indicate that increasing memory density and retrieval volume alone does not suffice to match the accuracy and efficiency of \ours.
\input{table/tab_density_budget.tex}
\input{table/tab_component_ablation.tex}
\subsection{Ablation Studies}
We ablate place-level memory scoping and the final evidence supplied to the \ac{LLM}, measuring accuracy and inference cost on \oursdata with the same content store and retrieval cues.

\textbf{Place Retrieval.}
Full-store search bypasses the place retrieval index and retrieves individual frame records directly from the entire content store (\cref{tab:component_ablation}).
\ours instead first selects \emph{places} to guide access to their associated records.
With place-level memory scoping, \ours is $2.4\times$ faster than full-store search and achieves slightly higher question accuracy.

\input{figure/fig_entrance_comparison.tex}
\textbf{Complementary Retrieval Signals.}
Place Synopsis-only retrieval ($\mathcal{I}_1$) accesses fewer frame records than the other variants and has the lowest latency, but its accuracy falls below full-store search.
Scene Codebook-only retrieval ($\mathcal{I}_2$) recovers much of this accuracy loss, but neither signal alone matches the accuracy of \ours.
With synopsis-first retrieval and codebook fallback, \ours accesses fewer frame records overall and has lower average query latency than codebook-only retrieval, while average token usage remains similar across retrieval variants.

\textbf{Final Evidence Construction.}
To examine the benefit of our evidence construction, \cref{tab:component_ablation} compares \ours with a ``Raw frame records'' variant.
We retain the upstream plans, retrieval, selection, and computations, but replace the final \ac{LLM} input with only the captions, positions, and timestamps of the same selected records, excluding computed evidence.
Providing plan-constructed evidence increases question accuracy from $72.9\%$ to $91.4\%$ at similar latency.
\subsection{Qualitative Results}
In \cref{fig:entrance_comparison}, we present two qualitative examples on \oursdata that require reasoning over observations from multiple visits.
Using retained observations, \ours identifies the different entrances used across visits (top) and confirms a prior encounter with the metal boat sculpture (bottom).
Note that the images visualize the retrieved frames, while the \ac{LLM} receives information from the corresponding frame records rather than the images themselves.

%% file: table/tab_evidence_plans.tex
\begin{table}[!t]
\centering
\caption{Mapping from evaluated query types and fine-grained intents to our three predefined evidence plans.}
\label{tab:evidence_plans}
\footnotesize
\setlength{\tabcolsep}{9pt}
\renewcommand{\arraystretch}{1}
\begin{tabular}{@{}ccc@{}}
    \toprule
    Query Type & Fine-Grained Intent & Type-Intent Evidence Plan \\
    \midrule
    Text & Descriptive \acs*{QA}
        & \multirow{3}{*}[-2.5pt]{Scene-Grounded Plan} \\
    \cmidrule{1-2}
    \multirow{3}{*}{Binary}
        & Ego-State                     & \\
        & Action and Relations & \\
        \cmidrule{3-3}
        & Object Existence
        & \multirow{2}{*}[-2.5pt]{Object-Grounded Plan} \\
    \cmidrule{1-2}
    \multirow{3}{*}{Position}
        & Object Localization & \\
    \cmidrule{3-3}
        & Scene Transition
        & \multirow{6}{*}[-5pt]{Trajectory-Grounded Plan} \\
        & Ego-Motion          & \\
    \cmidrule{1-2}
    \multirow{3}{*}{Time}
        & Sighting         & \\
        & Scene Transition & \\
        & Ego-Motion       & \\
    \cmidrule{1-2}
    Duration & Event-Endpoint Pair & \\
    \bottomrule
\end{tabular}
\end{table}

%% file: table/tab_navqa_results.tex
\begin{table*}[!t]
\centering
\setlength{\tabcolsep}{3.5pt}
\renewcommand{\arraystretch}{1.4}
\setlength{\aboverulesep}{0pt}
\setlength{\belowrulesep}{0pt}
\caption{Results on the \navqa dataset.
Results of \remembr and multi-frame \acs*{VLM} are partially taken from~\cite{anwar2025remembr,gorlo2026describe}.
${\dagger}$ indicates results obtained with in-context examples in the \acs*{LLM} prompts that overlap with the evaluation set.
}
\label{tab:navqa_results}
\noindent\resizebox{\textwidth}{!}{%
\begin{tabular}{clrggggccccgggg}
    \toprule
    \multicolumn{3}{r}{Metric:} & \multicolumn{4}{c}{\cellcolor{lightergray}Question Accuracy $\uparrow$} & \multicolumn{4}{c}{Positional Error [m] $\downarrow$} & \multicolumn{4}{c}{\cellcolor{lightergray}Temporal Error [min] $\downarrow$}\\
    \cmidrule(lr){4-7} \cmidrule(lr){8-11} \cmidrule(lr){12-15}
    \multicolumn{2}{l}{Method} & Memory Horizon & Short & Medium & Long & All & Short & Medium & Long & All & Short & Medium & Long & All\\
    \midrule
    \multirow{7}{*}{\rotatebox[origin=c]{90}{\remembr~\cite{anwar2025remembr}}}
    & \multicolumn{2}{l}{NVILA-Lite-2B + GPT-5-mini} & 0.423 & 0.519 & 0.524 & 0.483 & 11.179 & \snd{23.877} & 65.588 & 39.278 & 0.313 & \trd{1.240} & 4.472 & 2.518 \\
    & \multicolumn{2}{l}{NVILA-Lite-8B + GPT-5-mini} & 0.551 & \trd{0.605} & \trd{0.714} & 0.607 & 8.486 & 34.409 & 56.297 & 37.337 & 0.363 & 1.491 & 4.979 & 2.840 \\
    & \multicolumn{2}{l}{NVILA-Lite-8B + Qwen3.5-9B}${\dagger}$ & 0.423 & 0.481 & 0.571 & 0.478 & 13.155 & 49.443 & 66.114 & 48.461 & 0.685 & 1.822 & \trd{2.660} & \trd{1.961} \\
    & \multicolumn{2}{l}{NVILA-Lite-8B + GPT-5-mini}${\dagger}$ & 0.538 & \snd{0.630} & 0.571 & 0.582 & 8.467 & 53.279 & 53.717 & 41.485 & 0.436 & 1.336 & 4.122 & 2.414 \\
    & \multicolumn{2}{l}{NVILA-Lite-2B + GPT-5-mini}${\dagger}$ & 0.577 & 0.556 & 0.643 & 0.582 & 8.535 & \trd{25.030} & 49.259 & \trd{31.536} & \snd{0.272} & 1.248 & 4.975 & 2.743 \\
    & \multicolumn{2}{l}{VILA1.5-13B + GPT-4o}${\dagger}$ & \trd{0.62} & 0.58 & 0.65 & \trd{0.61} & \fst{5.1} & 27.5 & \trd{46.25} & \snd{29.96} & \trd{0.3} & 1.8 & 3.6 & 2.3 \\
    & \multicolumn{2}{l}{VILA1.5-13B + Llama3.1-8B}${\dagger}$ & 0.31 & 0.33 & 0.21 & 0.30 & 159.9 & 151.2 & 165.3 & 159.9 & 9.5 & 7.9 & 18.7 & 13.3 \\
    \arrayrulecolor{lightgray} \cmidrule(lr){1-15} \arrayrulecolor{black}
    \multicolumn{3}{l}{Multi-frame \acs*{VLM} | GPT-4o}${\dagger}$ & 0.55 & $\times$ & $\times$ & $\times$ & 7.5 & $\times$ & $\times$ & $\times$ & 0.5 & $\times$ & $\times$ & $\times$ \\
    \arrayrulecolor{lightgray} \cmidrule(lr){1-15} \arrayrulecolor{black}
    \multicolumn{3}{l}{\conceptgraphs~\cite{gu2024conceptgraphs} | NVILA-Lite-8B + GPT-5-mini } & 0.385 & 0.296 & 0.143 & 0.299 & 85.67 & 126.35 & 158.67 & 130.03 & $\times$ & $\times$ & $\times$ & $\times$ \\
    \arrayrulecolor{lightgray} \cmidrule(lr){1-15} \arrayrulecolor{black}
    \multicolumn{3}{l}{\daaam~\cite{gorlo2026describe} | DAM-3B + GPT-5-mini} & \snd{0.654} & \snd{0.630} & \snd{0.786} & \snd{0.672} & \trd{7.282} & 46.015 & \snd{42.116} & 33.890 & 0.443 & \fst{1.030} & \snd{2.538} & \snd{1.591} \\
    \arrayrulecolor{lightgray} \cmidrule(lr){1-15} \arrayrulecolor{black}

    \multicolumn{3}{l}{\emph{\acs*{RetrACER} (Ours) | NVILA-Lite-8B + Qwen3.5-9B}}
    & \fst{0.923} & \fst{0.741} & \fst{0.929} & \fst{0.851}
    & \snd{5.610} & \fst{12.254} & \fst{29.585} & \fst{18.126}
    & \fst{0.190} & \snd{1.081} & \fst{1.815} & \fst{1.198} \\
    \bottomrule
\end{tabular}%
}
\end{table*}

%% file: figure/fig_question_categories.tex
\begin{figure}[t]
    \centering
    \includegraphics[width=\linewidth]{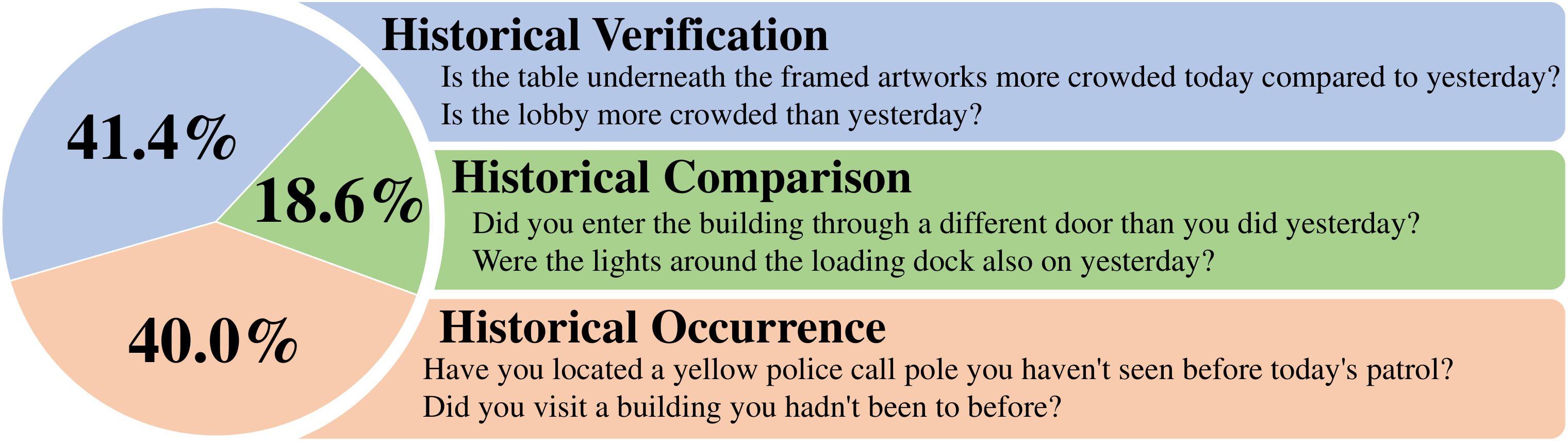}
    \caption{Question categories in \acs*{MS-NaVQA} and examples.}
    \label{fig:question_categories}
\end{figure}

%% file: table/tab_ocnavqa_results.tex
\begin{table}[!t]
\centering
\setlength{\tabcolsep}{4pt}
\renewcommand{\arraystretch}{1.4}
\setlength{\aboverulesep}{0pt}
\setlength{\belowrulesep}{0pt}
\caption{Results on the \ocnavqa dataset.}
\label{tab:ocnavqa_results}
\resizebox{\columnwidth}{!}{%
\begin{tabular}{lccc}
    \toprule
    Method
    & \makecell[l]{Question\\Accuracy $\uparrow$}
    & \makecell[l]{Positional\\Error [m] $\downarrow$}
    & \makecell[l]{Temporal\\Error [min] $\downarrow$} \\
    \midrule
    \remembr{} - NVILA-Lite-2B (GPT-5-mini)
    & 0.432
    & \trd{53.466}
    & \snd{2.287} \\
    \remembr{} - NVILA-Lite-8B (GPT-5-mini)
    & 0.463
    & 55.894
    & 4.106 \\
    \remembr{} - NVILA-Lite-8B (Qwen3.5-9B)
    & \trd{0.552}
    & 56.649
    & 4.196 \\
    \arrayrulecolor{lightgray}
    \cmidrule(l{.7em}){1-4}
    \arrayrulecolor{black}
    \conceptgraphs (GPT-5-mini)
    & 0.299
    & 111.29
    & $\times$ \\
    \arrayrulecolor{lightgray}
    \cmidrule(l{.7em}){1-4}
    \arrayrulecolor{black}
    \daaam (GPT-5-mini)
    & \snd{0.711}
    & \snd{41.75}
    & \fst{1.792} \\
    \daaam (Qwen3.5-9B)
    & 0.493
    & 54.44
    & 3.196 \\
    \arrayrulecolor{lightgray}
    \cmidrule(l{.7em}){1-4}
    \arrayrulecolor{black}
    \acs*{RetrACER} (Ours) - NVILA-Lite-8B (Qwen3.5-9B)
    & \fst{0.806}
    & \fst{38.555}
    & \trd{2.506} \\
    \bottomrule
\end{tabular}}
\end{table}

%% file: table/tab_msnavqa_results.tex
\begin{table}[!t]
\centering
\setlength{\tabcolsep}{4pt}
\renewcommand{\arraystretch}{1.4}
\setlength{\aboverulesep}{0pt}
\setlength{\belowrulesep}{0pt}
\caption{Results on \acs*{MS-NaVQA}. $\ddagger$ indicates our adaptation of \daaam for multi-session evaluation.}
\label{tab:msnavqa_results}
\resizebox{\columnwidth}{!}{%
\begin{tabular}{lccc}
    \toprule
    Method
    & \makecell[l]{Question\\Accuracy $\uparrow$}
    & \makecell[l]{Balanced\\Accuracy $\uparrow$}
    & \makecell[l]{Average\\Latency [s] $\downarrow$}\\
    \midrule
    Majority Pick & 0.629 & 0.500 & $\times$ \\
    \midrule
    \remembr{} - NVILA-Lite-8B (Qwen3.5-9B)
    & 0.600 & 0.603 & 61.520 \\
    \remembr{} - NVILA-Lite-8B (GPT-5-mini)
    & \trd{0.629} & \trd{0.681} & \trd{47.211} \\
    \arrayrulecolor{lightgray}
    \cmidrule(l{.7em}){1-4}
    \arrayrulecolor{black}
    \daaam{}\textsuperscript{$\ddagger$} (Qwen3.5-9B)
    & 0.529 & 0.570 & 86.504 \\
    \daaam{}\textsuperscript{$\ddagger$} (GPT-5-mini)
    & \snd{0.686} & \snd{0.711} & \snd{41.664} \\
    \arrayrulecolor{lightgray}
    \cmidrule(l{.7em}){1-4}
    \arrayrulecolor{black}
    \acs*{RetrACER} (Ours) - NVILA-Lite-8B (Qwen3.5-9B)
    & \fst{0.914} & \fst{0.908} & \fst{26.380}\\
    \bottomrule
\end{tabular}}
\end{table}

%% file: figure/fig_index_growth.tex
\begin{figure}[t]
    \centering
    \includegraphics[width=\linewidth]{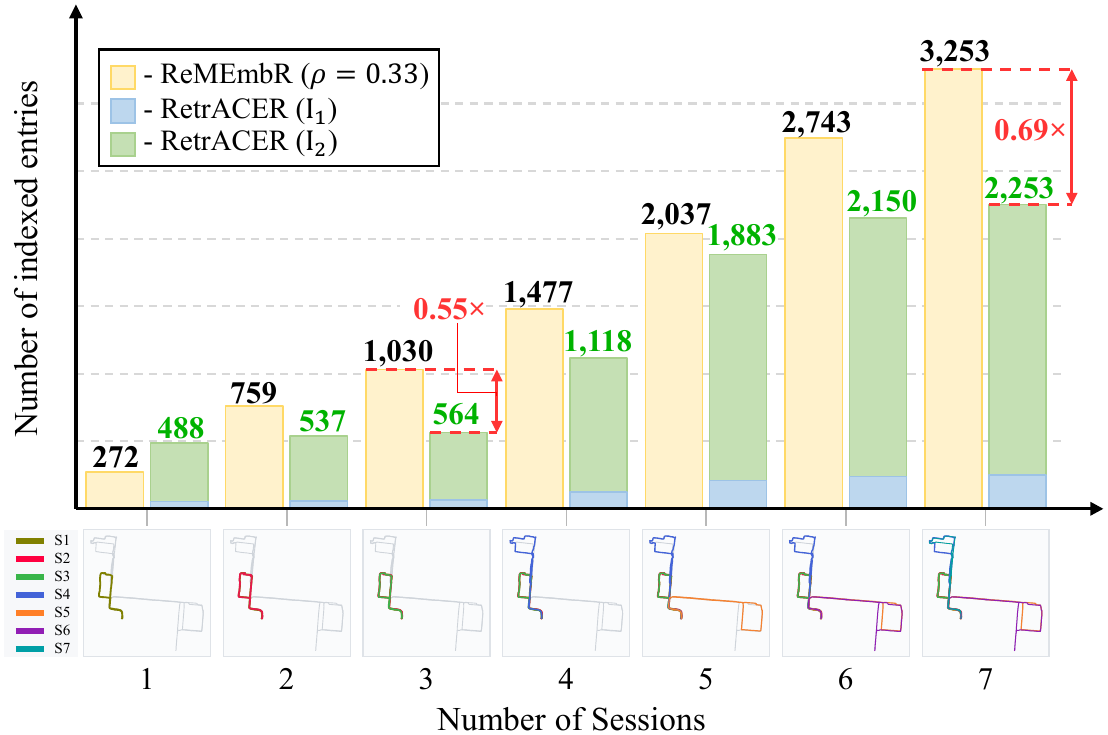}
    \caption{Retrieval-index growth across seven accumulated sessions.
    Yellow bars show \remembr, while stacked blue and green bars show the $\mathcal{I}_1$ and $\mathcal{I}_2$ of \acs*{RetrACER}, respectively.}
    \label{fig:index_growth}
\end{figure}

%% file: table/tab_density_budget.tex
\begin{table}[t]
\centering
\small
\sisetup{detect-weight=true, mode=text, group-digits=false}
\setlength{\tabcolsep}{2pt}
\renewcommand{\arraystretch}{1.15}

\caption{Performance and inference cost under different memory density and retrieval budget settings. $\rho$ denotes stored records per second, and $k$ denotes the number of records returned per tool call. Underlining indicates the default \remembr settings. All configurations use NVILA-Lite-8B and Qwen3.5-9B.}
\label{tab:density_budget}

\resizebox{\columnwidth}{!}{%
\begin{tabular}{@{}clc
    S[table-format=1.3]
    S[table-format=2.3]
    S[table-format=2.3]
    S[table-format=3.2]
    S[table-format=3.2]@{}}
\toprule
& \makecell{Method} & \makecell{$\rho / k$}
& {\makecell{Question\\Accuracy $\uparrow$}}
& {\makecell{Pos. Error $\downarrow$\\{[m]}}}
& {\makecell{Temp. Error $\downarrow$\\{[min]}}}
& {\makecell{Avg. Latency $\downarrow$\\{[s]}}}
& {\makecell{Avg. Tokens $\downarrow$\\{[K]}}} \\
\midrule

\multirow{4}{*}[-0.47ex]{\rotatebox[origin=c]{90}{\navqa}}
& \multirow{3}{*}{\remembr{}~\cite{anwar2025remembr}} & \underline{0.33 / 5}
& 0.478 & \cellcolor{third}48.461 & \cellcolor{second}1.961 & \cellcolor{third}48.75 & \cellcolor{third}22.53 \\
& & 10 / 5
& \cellcolor{second}0.597 & \cellcolor{second}37.081 & \cellcolor{third}2.194 & \cellcolor{second}48.05 & \cellcolor{second}18.54 \\
& & 10 / 150
& \cellcolor{third}0.582 & 65.300 & 6.093 & 184.04 & 226.99 \\
\arrayrulecolor{lightgray}
\cmidrule(l{.7em}){2-8}
\arrayrulecolor{black}
& \acs*{RetrACER} & 10 / $\times$
& \cellcolor{first}\bfseries 0.851
& \cellcolor{first}\bfseries 18.126
& \cellcolor{first}\bfseries 1.198
& \cellcolor{first}\bfseries 27.65
& \cellcolor{first}\bfseries 8.91 \\
\midrule

\multirow{4}{*}[-0.47ex]{\rotatebox[origin=c]{90}{\ocnavqa}}
& \multirow{3}{*}{\remembr{}~\cite{anwar2025remembr}} & \underline{0.33 / 5}
& 0.552 & \cellcolor{third}56.649 & \cellcolor{second}4.196 & \cellcolor{third}63.31 & \cellcolor{third}35.56 \\
& & 10 / 5
& \cellcolor{third}0.612 & 62.434 & \cellcolor{third}4.237 & \cellcolor{second}56.99 & \cellcolor{second}28.34 \\
& & 10 / 150
& \cellcolor{second}0.672 & \cellcolor{second}50.153 & 17.196 & 230.32 & 255.21 \\
\arrayrulecolor{lightgray}
\cmidrule(l{.7em}){2-8}
\arrayrulecolor{black}
& \acs*{RetrACER} & 10 / $\times$
& \cellcolor{first}\bfseries 0.806
& \cellcolor{first}\bfseries 38.555
& \cellcolor{first}\bfseries 2.506
& \cellcolor{first}\bfseries 28.39
& \cellcolor{first}\bfseries 9.26 \\
\midrule

\multirow{4}{*}[-0.47ex]{\rotatebox[origin=c]{90}{\acs*{MS-NaVQA}}}
& \multirow{3}{*}{\remembr{}~\cite{anwar2025remembr}} & \underline{0.33 / 5}
& \cellcolor{second}0.600 & {$\times$} & {$\times$} & \cellcolor{third}61.52 & \cellcolor{third}28.75 \\
& & 10 / 5
& 0.443 & {$\times$} & {$\times$} & \cellcolor{second}56.05 & \cellcolor{second}24.45 \\
& & 10 / 150
& \cellcolor{third}0.486 & {$\times$} & {$\times$} & 163.38 & 219.70 \\
\arrayrulecolor{lightgray}
\cmidrule(l{.7em}){2-8}
\arrayrulecolor{black}
& \acs*{RetrACER} & 10 / $\times$
& \cellcolor{first}\bfseries 0.914
& {$\times$}
& {$\times$}
& \cellcolor{first}\bfseries 26.38
& \cellcolor{first}\bfseries 4.82 \\
\bottomrule
\end{tabular}%
}
\end{table}

%% file: table/tab_component_ablation.tex
\begin{table}[!t]
\centering
\setlength{\tabcolsep}{5pt}
\renewcommand{\arraystretch}{1.25}
\setlength{\aboverulesep}{0pt}
\setlength{\belowrulesep}{0pt}
\caption{Ablation study of \emph{\acs*{RetrACER}} on \acs*{MS-NaVQA}.}
\label{tab:component_ablation}
\resizebox{\columnwidth}{!}{%
\begin{tabular}{lcccc}
    \toprule
    Method
    & \makecell[l]{Question\\Accuracy $\uparrow$}
    & \makecell[l]{Balanced\\Accuracy $\uparrow$}
    & \makecell[l]{Average\\Latency [s] $\downarrow$}
    & \makecell[l]{Average\\Tokens [K] $\downarrow$}\\
    \midrule
    Place Synopsis only ($I_1$) & 0.686 & 0.719 & \fst{12.53} & 4.85 \\
    Scene Codebook only ($I_2$) & \trd{0.871} & \trd{0.858} & 28.89 & \fst{4.81} \\
    Full-store search & \snd{0.886} & \snd{0.885} & 63.68 & \trd{4.83} \\
    \arrayrulecolor{lightgray}
    \cmidrule(l{.7em}){1-5}
    \arrayrulecolor{black}
    Raw frame records & 0.729 & 0.721 & \trd{26.77} & 5.09 \\
    \midrule
    \textbf{\acs*{RetrACER} (full)} & \fst{0.914} & \fst{0.908} & \snd{26.38} & \snd{4.82} \\
    \bottomrule
\end{tabular}}
\end{table}

%% file: figure/fig_entrance_comparison.tex
\begin{figure*}[t!]
    \centering
    \includegraphics[width=0.91\textwidth]{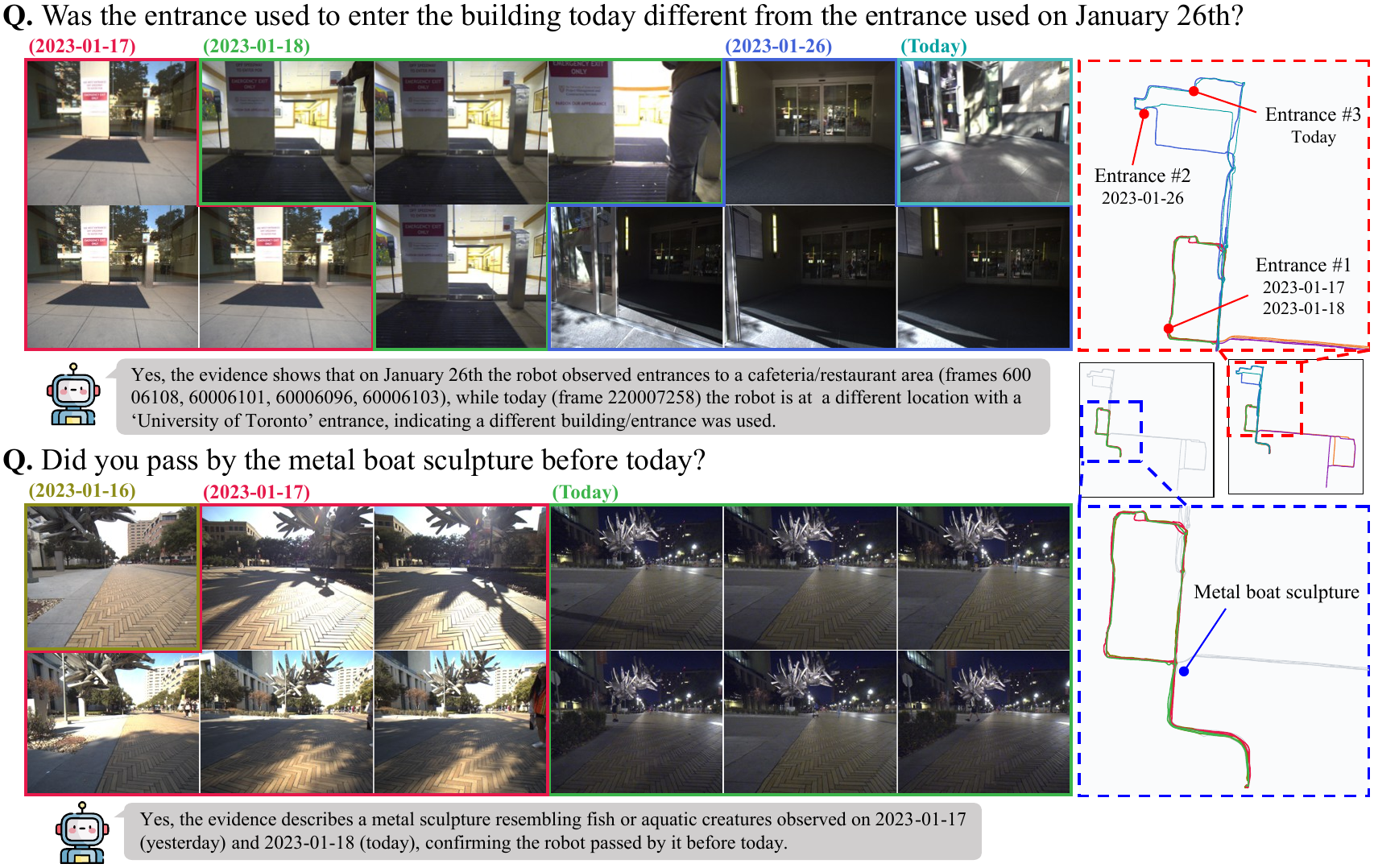}
    \caption{Qualitative examples of cross-session question answering on \acs*{MS-NaVQA}: entrance comparison (top) and verification of a prior encounter with a metal boat sculpture (bottom).
    Retrieved frames and trajectory visualizations are shown alongside the actual responses of \acs*{RetrACER}.
    The reference to ``University of Toronto'' in the top response is inherited from a captioner hallucination.}
    \label{fig:entrance_comparison}
    \vspace{-5mm}
\end{figure*}

%% file: sec/limitations.tex
\section{Limitations and Future Work}
\label{sec:limitations}
Several limitations of the current system remain to be addressed.
First, \textbf{Frame Caption Errors}. For example, captions in our experiments described a moving car as stationary and a person opening a door as standing beside it.
Captions may also omit small objects or introduce unsupported details.
Once stored, these errors can propagate to subsequent answers, as illustrated by the hallucinated institution name in \cref{fig:entrance_comparison}.
Second, \textbf{Place Content Consolidation}. While the place retrieval index of \ourmem is bounded by spatial coverage for fixed $\ell$ and $K$, the content store continues to grow as observations accumulate.
Consolidating redundant observations within each place while preserving query-relevant historical evidence may reduce storage requirements over longer deployments.
Third, \textbf{Object Re-identification}. Object association relies on spatial proximity and visual similarity, without a dedicated mechanism for cross-session re-identification.
Consequently, nearby instances with similar appearance may be merged, whereas the same object observed at different locations may be stored as separate records.
Incorporating cross-session re-identification, as in \ltmem~\cite{lee2026lt}, may improve object identity consistency across visits.

%% file: sec/conclusion.tex
\section{Conclusion}
\label{sec:conclusion}
We present \ours, a multi-session navigation memory system for efficient lifelong spatio-temporal reasoning.
By decoupling the place retrieval index from the content store, \ourmem bounds index growth by spatial coverage for fixed $\ell$ and $K$, while preserving detailed object records and frame records.
Using Type-Intent Evidence Plans, \ours constructs question-specific evidence from these records without iterative \ac{LLM}-directed tool calls.
With a locally hosted 9B model, \ours achieves the highest question accuracy among the evaluated methods on two single-session datasets and \oursdata, where it also has the lowest query latency.
Taken together, our findings highlight the importance of organizing both access to accumulated navigation experience and its transformation into question-specific evidence.
Place-level memory scoping and record-level intent-guided evidence construction jointly support effective and efficient reasoning, with \ourmem and our three evidence plans providing one concrete realization for the evaluated navigation questions.

%% file: sec/supplementary.tex
\clearpage
\onecolumn
\acuseall
\setcounter{section}{0}
\setcounter{table}{0}
\setcounter{figure}{0}
\renewcommand{\thesection}{S\arabic{section}}
\renewcommand{\thetable}{S\arabic{table}}
\renewcommand{\thefigure}{S\arabic{figure}}
\renewcommand{\theHsection}{S\arabic{section}}
\renewcommand{\theHtable}{S\arabic{table}}
\renewcommand{\theHfigure}{S\arabic{figure}}
\begin{center}
\Large\textbf{\thetitle}\\
\vspace{0.5em}Supplementary Material\\
\vspace{1.0em}
\end{center}
\section{MS-NaVQA Dataset Details}
\label{sec:dataset_details}
Section~\ref{sec:dataset_taxonomy} describes how \oursdata was constructed and how its questions are categorized, and Section~\ref{sec:inventory} lists all questions.
\input{sec/supplementary_dataset}
\subsection{Complete Question Inventory}
\label{sec:inventory}
Tables~\ref{tab:inventory_verification}--\ref{tab:inventory_occurrence} list all 70 questions by category, sorted by session; the session of a question equals its Number of Sessions in Table~\ref{tab:session_context}.
Following \navqa~\cite{anwar2025remembr} and \ocnavqa~\cite{gorlo2026describe}, each question is preceded by the current time and the robot's position, \eg, ``The current time is 2023-01-17 14:25:37 and you are located at [0.4, -0.09, -0.2].''
The tables omit this prefix for brevity, and Table~\ref{tab:session_context} lists the time and position of each session.
GT denotes the ground-truth answer.
\input{table/tab_question_inventory}

%% file: sec/supplementary_dataset.tex
\subsection{Construction and Categories}
\label{sec:dataset_taxonomy}
\paragraph{Construction.}
\oursdata consists of 70 binary questions for evaluating reasoning over navigation experience accumulated across sessions, each of which is one of the CODa~\cite{zhang2024robust} sequences used in prior work~\cite{anwar2025remembr,gorlo2026describe}.
Whereas prior work evaluates each of these sequences (0, 3, 4, 6, 16, 21, and 22) as a single session, we accumulate them in chronological order as Sessions 1--7.
Session~1 provides the initial observation history, and each question is asked near the end of one of the remaining six sessions (Table~\ref{tab:session_context}), using the memory built from that session and all preceding ones.
We designed the questions and annotated their ground-truth answers by inspecting the RGB observations of multiple sessions, so that answering them involves the observation history accumulated over sessions.

\input{table/tab_question_taxonomy}
\input{table/tab_session_context}
\paragraph{Category assignment.}
We assign each question to exactly one of the three categories in Table~\ref{tab:taxonomy}, based on the operation it requests rather than on its wording.
Verification questions are anchored at the current observation, whereas Comparison questions are anchored at a specified earlier time or visit.
For instance, ``Is the bicycle next to the green container still there today?'' is Verification, while ``Was the bicycle \ldots{} also there on January 26th like today?'' is Comparison.

%% file: table/tab_question_taxonomy.tex
\begin{table}[H]
\centering\small
\setlength{\tabcolsep}{5pt}
\renewcommand{\arraystretch}{1.15}
\caption{MS-NaVQA question categories, listed without the prefix ``Historical.'' Each question has exactly one category.}
\label{tab:taxonomy}
\begin{tabularx}{\textwidth}{@{}l X r@{}}
\toprule
Category & Definition & Count \\
\midrule
Verification & Change in a currently observed condition (quantity, crowding, presence, or location) from earlier observations. & 29 \\
Comparison & An event, state, or identity at a specified earlier time or visit, compared with another observation. & 13 \\
Occurrence & Prior observation or first encounter with an object, place, or event in the robot's observation history. & 28 \\
\bottomrule
\end{tabularx}
\end{table}

%% file: table/tab_session_context.tex
\begin{table}[H]
\centering\small
\caption{Accumulated sessions (including the current one), current time, and robot position for the questions of each session.}
\label{tab:session_context}
\begin{tabular}{clrlr}
\toprule
Number of Sessions & CODa seq. & Current time & Position [m] & Questions \\
\midrule
2 & (0, 3) & 2023-01-17 14:25:37 & [0.4, -0.09, -0.2] & 16 \\
3 & (0, 3, 4) & 2023-01-18 18:35:08 & [1.42, 7.81, -0.24] & 12 \\
4 & (0, 3, 4, 6) & 2023-01-26 14:44:56 & [1.45, 1.73, -0.21] & 12 \\
5 & (0, 3, 4, 6, 16) & 2023-02-08 13:02:35 & [0.84, 8.56, -0.46] & 12 \\
6 & (0, 3, 4, 6, 16, 21) & 2023-02-10 12:37:32 & [-0.58, -0.30, -0.11] & 10 \\
7 & (0, 3, 4, 6, 16, 21, 22) & 2023-02-10 15:47:50 & [0.75, -0.47, 0.03] & 8 \\
\bottomrule
\end{tabular}
\end{table}

%% file: table/tab_question_inventory.tex
\begingroup\small
\setlength{\tabcolsep}{5pt}
\renewcommand{\arraystretch}{1.1}
\begin{longtable}{@{}c p{0.82\textwidth} c@{}}
\caption{Historical Verification: all 29 questions.}\label{tab:inventory_verification}\\
\toprule Session & Question & GT \\ \midrule
\endfirsthead
\multicolumn{3}{l}{\textit{Historical Verification (continued)}}\\
\toprule Session & Question & GT \\ \midrule
\endhead
\midrule\multicolumn{3}{r}{\textit{Continued on next page}}\\
\endfoot
\bottomrule\endlastfoot
2 & Does the bike rack next to the gray concrete wall hold more bicycles today than on your last patrol? & yes \\
2 & Were there more vehicles at the loading zone than yesterday? & yes \\
2 & Is the lobby more crowded than yesterday? & yes \\
2 & Were the red benches and tables in the atrium more crowded today than yesterday? & yes \\
2 & Was the yellow sign next to the large tree outside the building lying down on the previous visit? & yes \\
2 & Are there more people around the loading dock than yesterday? & yes \\
2 & Is the brick-paved walkway between the buildings more crowded than on January 16th? & yes \\
2 & Is the table underneath the framed artworks more crowded today compared to yesterday? & yes \\
2 & Is the A-frame sign with ``Lucky Lab'' written on it, which was next to the large tree yesterday, still there today? & no \\
\arrayrulecolor{lightgray}\cmidrule(l{.7em}){1-3}\arrayrulecolor{black}
3 & Was the blue trash bin in the hallway at the same place yesterday as today? & yes \\
3 & Did you see a person with a lifevest today? & no \\
3 & Is the bicycle that was next to the green container yesterday still there today? & no \\
3 & Have the trucks that appeared at the loading dock yesterday disappeared today? & yes \\
\arrayrulecolor{lightgray}\cmidrule(l{.7em}){1-3}\arrayrulecolor{black}
4 & Did the number of trucks at the loading zone increase since your last patrol? & yes \\
4 & Has the blue truck moved since you last saw it? & yes \\
4 & Is the ``Texas'' white van that was parked next to the hydrant still there? & no \\
\arrayrulecolor{lightgray}\cmidrule(l{.7em}){1-3}\arrayrulecolor{black}
5 & Is the black truck next to the green container still there today? & no \\
5 & Are people carrying umbrellas on the brick walkway today, unlike your previous visit? & yes \\
5 & Are there more vehicles inside the loading dock than on your previous visit? & no \\
\arrayrulecolor{lightgray}\cmidrule(l{.7em}){1-3}\arrayrulecolor{black}
6 & Is the orangish utility cart at the loading dock gone today? & yes \\
6 & Is the ``PTS IS HIRING'' sign near the intersection still there today? & yes \\
6 & Are there fewer vehicles parked inside the loading dock than on your previous visit? & yes \\
6 & Is the CVS Pharmacy still there today? & yes \\
6 & Is the bicycle next to the green container still there today? & yes \\
6 & Are there more white delivery trucks near the loading dock than on your previous visit? & yes \\
\arrayrulecolor{lightgray}\cmidrule(l{.7em}){1-3}\arrayrulecolor{black}
7 & Are there more white delivery trucks near the loading dock than earlier today? & no \\
7 & Is the bicycle next to the green container that you saw earlier today still there? & yes \\
7 & Is the Longhorn Coffee Co. more crowded today than on your last patrol? & yes \\
7 & Is the yellow caution sign near Longhorn Coffee Co. still there today? & no \\
\end{longtable}
\begin{longtable}{@{}c p{0.82\textwidth} c@{}}
\caption{Historical Comparison: all 13 questions.}\label{tab:inventory_comparison}\\
\toprule Session & Question & GT \\ \midrule
\endfirsthead
\multicolumn{3}{l}{\textit{Historical Comparison (continued)}}\\
\toprule Session & Question & GT \\ \midrule
\endhead
\midrule\multicolumn{3}{r}{\textit{Continued on next page}}\\
\endfoot
\bottomrule\endlastfoot
2 & Did the same person open the door for you as last time? & no \\
2 & Was that white van at the loading dock there yesterday too? & no \\
2 & Was the bicycle parked next to the green container also there on January 16th? & no \\
2 & Did you see the same yellow police call pole as yesterday? & yes \\
2 & Were the bicycles next to the green container there on the previous day as well? & no \\
\arrayrulecolor{lightgray}\cmidrule(l{.7em}){1-3}\arrayrulecolor{black}
3 & Did you see the horse sculpture on January 16th, January 17th, and also today? & no \\
3 & Were the lights around the loading dock also on yesterday? & no \\
3 & Was the loading dock empty two days ago, as it is today? & yes \\
3 & Did you enter the building through a different door than you did yesterday? & yes \\
\arrayrulecolor{lightgray}\cmidrule(l{.7em}){1-3}\arrayrulecolor{black}
4 & Was the bicycle that is next to the black truck today not there on January 18th? & yes \\
\arrayrulecolor{lightgray}\cmidrule(l{.7em}){1-3}\arrayrulecolor{black}
5 & Was the bicycle that is next to the green container also there on January 26th like today? & yes \\
\arrayrulecolor{lightgray}\cmidrule(l{.7em}){1-3}\arrayrulecolor{black}
7 & Were the Valentine's Day decorations near the restroom area also there the last time you visited this area? & no \\
7 & Was the entrance used to enter the building today different from the entrance used on January 26th? & yes \\
\end{longtable}
\begin{longtable}{@{}c p{0.82\textwidth} c@{}}
\caption{Historical Occurrence: all 28 questions.}\label{tab:inventory_occurrence}\\
\toprule Session & Question & GT \\ \midrule
\endfirsthead
\multicolumn{3}{l}{\textit{Historical Occurrence (continued)}}\\
\toprule Session & Question & GT \\ \midrule
\endhead
\midrule\multicolumn{3}{r}{\textit{Continued on next page}}\\
\endfoot
\bottomrule\endlastfoot
2 & Was the blue truck absent on every patrol before today? & yes \\
2 & Did you see the white van with ``Texas'' written on it for the first time today, not on any previous day? & yes \\
\arrayrulecolor{lightgray}\cmidrule(l{.7em}){1-3}\arrayrulecolor{black}
3 & Was the Wet Floor sign next to the round column at the corridor there before? & no \\
3 & Have you patrolled the campus at night before today? & no \\
3 & Was the ``EMERGENCY EXIT ONLY'' marker there before? & yes \\
3 & Did you pass by the metal boat sculpture before today? & yes \\
\arrayrulecolor{lightgray}\cmidrule(l{.7em}){1-3}\arrayrulecolor{black}
4 & Is today the first time you've seen a black truck next to the green container? & no \\
4 & Was the Wetzel's Pretzels food truck there before today? & no \\
4 & Did you visit a building you hadn't been to before? & yes \\
4 & Is this the first time you have seen the white truck parked in the loading dock? & no \\
4 & Did you see the black truck with ``COTTON'' written on the back for the first time? & yes \\
4 & Did you see the restroom today for the first time? & yes \\
4 & Did you see a person pulling a wagon along the brick-paved walkway other than today? & no \\
4 & Have you seen a yellow truck with ``PENSKE'' written on it other than today? & no \\
\arrayrulecolor{lightgray}\cmidrule(l{.7em}){1-3}\arrayrulecolor{black}
5 & Is it the first time you have seen the Wetzel's Pretzels food truck? & no \\
5 & Did you see the rainbow flag for the first time today? & yes \\
5 & Was the orangish utility cart at the loading dock visible only today? & yes \\
5 & Did you see ``Wells Fargo'' for the first time on today's patrol? & yes \\
5 & Did you see the red semi-truck for the first time today, not on any previous day? & yes \\
5 & Have you never seen the ``CHURCH PARKING ONLY'' sign on any patrol before today? & yes \\
5 & Have you located a crosswalk before today? & no \\
5 & Did you observe the CVS Pharmacy today for the first time? & yes \\
\arrayrulecolor{lightgray}\cmidrule(l{.7em}){1-3}\arrayrulecolor{black}
6 & Have you seen ``Wells Fargo'' before? & yes \\
6 & Have you seen a new intersection you've never seen before? & yes \\
6 & Have you located a crosswalk before today? & yes \\
6 & Have you located a yellow police call pole you haven't seen before today's patrol? & yes \\
\arrayrulecolor{lightgray}\cmidrule(l{.7em}){1-3}\arrayrulecolor{black}
7 & Is this your first time visiting Longhorn Coffee Co.? & no \\
7 & Have you observed a van with ``OTIS'' written on it before? & no \\
\end{longtable}
\endgroup